\documentclass[10pt,twocolumn,letterpaper]{article}

\usepackage[pagenumbers]{cvpr} 

\usepackage{graphicx}
\usepackage{amsmath}
\usepackage{amssymb}
\usepackage{booktabs}
\usepackage{soul}
\usepackage{multirow}

\makeatletter
\@namedef{ver@everyshi.sty}{}
\makeatother

\usepackage{tikz}
\usepackage{pgfplots}
\pgfplotsset{compat=1.17}

\usepackage[pagebackref,breaklinks,colorlinks]{hyperref}

\usepackage[capitalize]{cleveref}
\crefname{section}{Sec.}{Secs.}
\Crefname{section}{Section}{Sections}
\Crefname{table}{Table}{Tables}
\crefname{table}{Tab.}{Tabs.}

\def\confName{CVPR}

\begin{document}

\title{STMR: Spiral Transformer for Hand Mesh Reconstruction}


\author{
Huilong Xie$^{1,2}$, Wenwei Song$^{1,2}$, Wenxiong Kang$^{1,2}$\thanks{Corresponding author.}, Yihong Lin$^{1,2}$
\\
$^{1}$South China University of Technology, Guangzhou, China,
$^{2}$Pazhou Lab, Guangzhou, China,\\
\tt\small{\{auhlxie, ausongwenwei, 202221017671\}@mail.scut.edu.cn, auwxkang@scut.edu.cn}
}

\maketitle

\begin{abstract}
Recent advancements in both transformer-based methods and spiral neighbor sampling techniques have greatly enhanced hand mesh reconstruction. Transformers excel in capturing complex vertex relationships, and spiral neighbor sampling is vital for utilizing topological structures. This paper ingeniously integrates spiral sampling into the Transformer architecture, enhancing its ability to leverage mesh topology for superior performance in hand mesh reconstruction, resulting in substantial accuracy boosts. STMR employs a single image encoder for model efficiency. To augment its information extraction capability, we design the multi-scale pose feature extraction (MSPFE) module, which facilitates the extraction of rich pose features, ultimately enhancing the model's performance. Moreover, the proposed predefined pose-to-vertex lifting (PPVL) method improves vertex feature representation, further boosting reconstruction performance. Extensive experiments on the FreiHAND dataset demonstrate the state-of-the-art performance and unparalleled inference speed of STMR compared with similar backbone methods, showcasing its efficiency and effectiveness. The code is available at \url{https://github.com/SmallXieGithub/STMR}.
\end{abstract}

\section{Introduction}
Reconstructing 3D hand mesh from a monocular image is a popular research topic, due to its extensive applications in virtual reality \cite{han2020megatrack}, behavior understanding \cite{joo2019towards}, \emph{etc}. Nevertheless, this task poses significant challenges, particularly owing to the inherent limitations imposed by a single-view perspective.

\begin{figure}[h]
\centering
\includegraphics[scale=0.46]{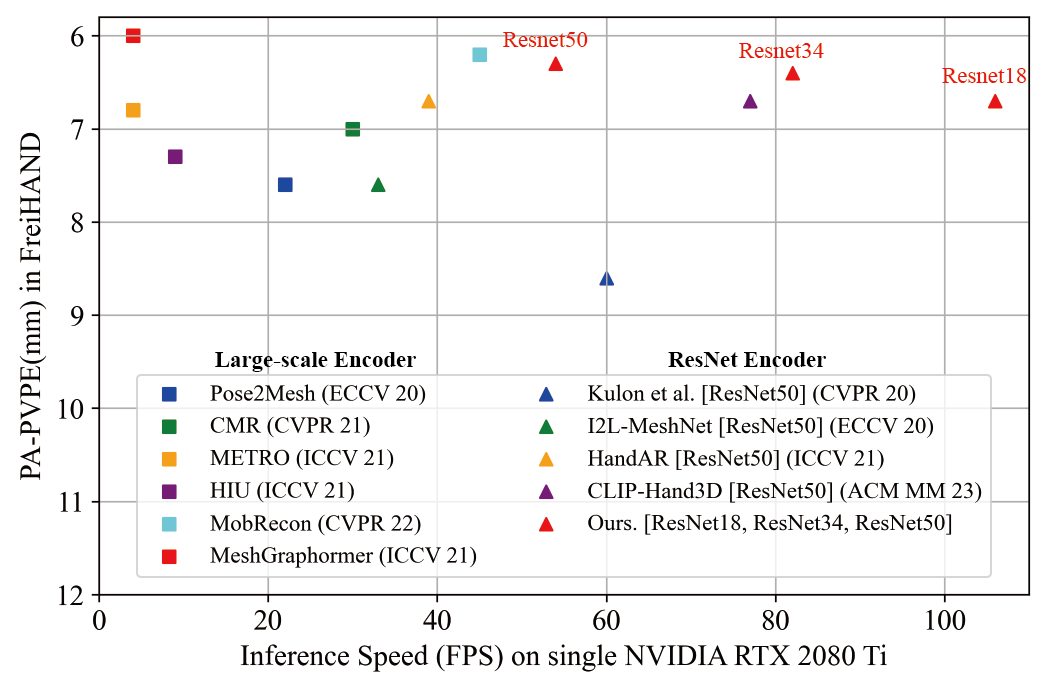}
\caption{Accuracy vs. inference speed on the FreiHAND test set. The proposed method outperforms competing techniques with a similar-scale visual encoder, demonstrating superior speed and performance. Best viewed in color.}
\label{fig_PV_FPS}
\end{figure}

Recently, vertex spiral neighbor-based modeling methods have made significant advances in mesh reconstruction. For example, \cite{gong2019spiralnet++} had suggested using fully connected layers to regress 3D mesh coordinates by explicitly formulating the order of aggregating neighbor vertices. A recent study \cite{chen2021camera} enhanced the inter-vertex relationships by utilizing a decoder with multi-scale spiral neighbors. Subsequently, \cite{chen2022mobrecon} further improved the efficiency of the 3D decoder by replacing the fully connected layer with deep-wise convolution. However, as discussed in the literature \cite{wu2020visual}, convolutional and fully connected layers have limitations in modeling long-range relationships. In contrast, Transformers are better at capturing long-term relationships.

Transformers also have achieved significant advances in hand mesh reconstruction. For instance, \cite{lin2021end} proposed Transformer-based structures for 3D hand mesh estimation. However, their approaches do not utilize the mesh topology, resulting in sub-optimal outcomes. Subsequently, Graph Convolutional Networks (GCNs) were employed to incorporate structural information into Transformer, further enhancing mesh reconstruction \cite{lin2021mesh}. Nonetheless, the integration of GCNs increases computational complexity, affecting model efficiency. A recent work \cite{guo2023clip} employed Transformers to regress the spatial coordinates of vertices on a predefined coarse-to-fine mesh, but it struggled to guide the network in effectively utilizing mesh topology through position encoding. In contrast, spiral neighbor sampling explicitly formulates the order of aggregating neighbor vertices, which contains the mesh topology between the center vertex and surrounding vertices and helps guide the network in revealing the relationships among mesh vertices. The above analysis motivates us to combine the spiral neighbor sample and Transformers for the 3D reconstruction of hand mesh and explore how spiral neighbor sampling can effectively inject mesh topology into the Transformer.

In this paper, we present a Spiral Transformer for reconstructing hand mesh from a monocular image. We use a Transformer to process unfolded spiral neighbor vertex features, enhancing vertices' interaction with their surroundings. At the same time, we introduce the topological order of the mesh to the Transformer through spiral neighbor sampling, which provides structural information to the 3D decoder. The Spiral Transformer optimizes 3D coordinate prediction by focusing on neighbor vertices. Our framework is organized in three steps: 2D encoding, 2D to 3D mapping, and 3D decoding. To enhance image feature extraction, we design a multi-scale pose feature extraction (MSPFE) module to obtain rich pose information through multi-layer feature sampling and fusion. In addition, we propose a predefined pose-to-vertex lifting (PPVL) method to enhance the accuracy of the vertex features by utilizing the prior information of the MANO model. As shown in Figure \ref{fig_PV_FPS}, we achieve better performance in terms of accuracy and speed compared with methods that employ comparable backbone architectures, demonstrating its efficiency and effectiveness.

Our main contributions are summarized as follows:

\begin{itemize}
\item We present a novel Spiral Transformer for reconstructing hand mesh, which enhances the 3D decoder to regress on 3D vertex coordinates by explicitly formulating the order of aggregating neighbor vertices.
\item We design the MSPFE module and PPVL for extracting pose and vertex features, respectively, which help estimate the 3D mesh vertex coordinates more accurately.
\item We demonstrate that our method achieves superior performance in terms of model efficiency and reconstruction accuracy via comprehensive evaluations and comparisons with state-of-the-art approaches that employ comparable backbone architectures.
\end{itemize}

\section{Related Work}
\subsection{Hand Mesh Reconstruction}
Hand mesh reconstruction has experienced significant advancements in recent years, driven by deep learning. Learning-based methods have attracted substantial interest among researchers, and there are two prevalent approaches: parametric and non-parametric approaches. The parametric methods involve estimating the pose and shape parameters of the MANO model \cite{MANO:SIGGRAPHASIA:2017} to generate a hand mesh. The MANO model's differentiability enables end-to-end optimization from single-view images \cite{boukhayma20193d, zhang2019end, zimmermann2019freihand, pavlakos2023reconstructing}. However, the nonlinearity of parameter estimation poses a challenge, particularly when attempting to achieve precise mesh reconstruction from a small set of parameters. Instead of regressing the model parameters, non-parametric approaches directly regress hand mesh, which usually follows three stages: 2D embedding, 2D-to-3D mapping, and 3D decoding \cite{ge20193d, kulon2020weakly, chen2021camera, chen2022mobrecon, guo2023clip}. 

In this paper, we propose a non-parametric approach called STMR. With the designed MSPFE module and PPVL, the network can obtain rich pose features and better model the mapping relationship between the 2D image and the 3D mesh, resulting in a more accurate hand mesh reconstruction.

\subsection{Transformer-based Reconstruction Model}
Hand mesh reconstruction is a complex task that not only needs to consider the relationship between image features and vertex features, but also needs to model the correlation among mesh vertices. Generally, convolutional and fully connected layers struggle to capture distant dependencies among vertices effectively, whereas Transformers excel at modeling long-range positional relationships. The Transformer-based models have been successfully used in mesh reconstruction works \cite{lin2021mesh, lin2021end, guo2023clip, pavlakos2024reconstructing}. Lin \emph{et al.} successively proposed Transformer-based structures \cite{lin2021end} and a carefully designed Mesh-Graphormer \cite{lin2021mesh} for 3D hand mesh estimation. Guo \emph{et al.} designed a Transformer mesh regressor to locate the spatial positional encodings among all sparse-to-dense mesh vertices. Pavlakos \emph{et al.} used ViT to estimate the model parameters of MANO from a monocular image.

In this paper, we exploit the mesh's topological information through the spiral serialization property of neighbor vertices \cite{gong2019spiralnet++} and combine it with Transformer to model the correlation among vertex features.

\section{Method}

\begin{figure*}[h]
\centering
\includegraphics[scale=0.6]{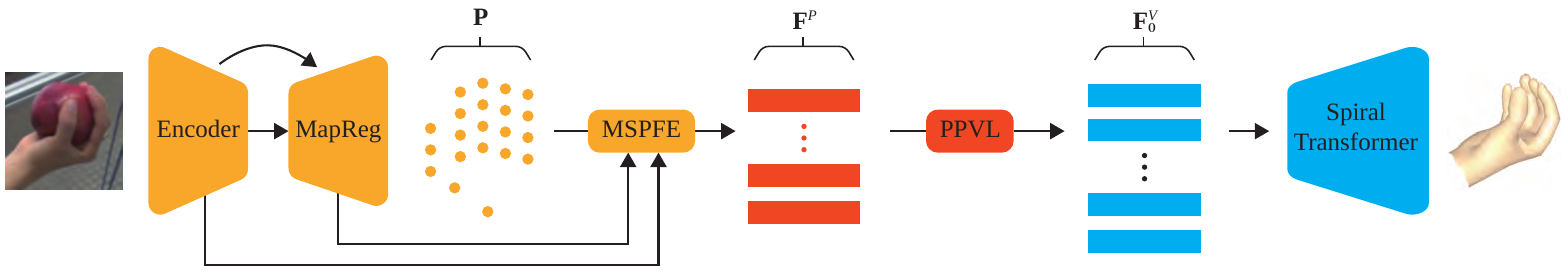}
\caption{Overview of our STMR framework. Best viewed in color.}
\label{fig_overall}
\end{figure*}

In this work, our goal is to estimate a 3D hand mesh with predicted vertices from a monocular image. Figure \ref{fig_overall} illustrates the overall framework of STMR. In this section, we will introduce its various sub-modules.

\begin{figure}[h]
\centering
\includegraphics[scale=1]{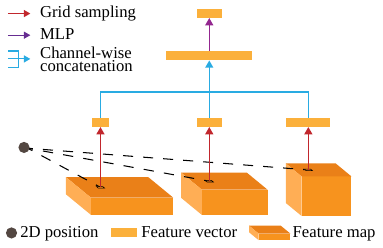}
\caption{Details of MSPFE module. Here, we use three feature maps at different scales as an example. Best viewed in color.}
\label{fig_mspfe}
\end{figure}

\subsection{Pose Feature Extraction}
\paragraph{Visual Encoder.} Following previous works \cite{hasson2019learning, chen2021camera, guo2023clip}, we employ ResNet \cite{he2016deep} as the Visual Encoder to encode the input monocular RGB image. Given the input monocular RGB image  $\mathbf{I} \in \mathbb{R}^{(B,3,H,W)}$, visual encoder outputs multi-level image features $\mathbf{F}^E_0 \in \mathbb{R}^{(B,C_0,H/2,W/2)}$, $\mathbf{F}^E_1 \in \mathbb{R}^{(B,C_1,H/4,W/4)}$, $\mathbf{F}^E_2 \in \mathbb{R}^{(B,C_2,H/8,W/8)}$, $\mathbf{F}^E_3 \in \mathbb{R}^{(B,C_3,H/16,W/16)}$ and $\mathbf{F}^E_4 \in \mathbb{R}^{(B,C_4,H/32,W/32)}$. Multi-level image features can provide global-local information. They are fed to the MSPFE module to obtain rich pose features. $\mathbf{F}^E_3$ and $\mathbf{F}^E_4$ are fed to 2D joint regressor to estimate 2D hand joint positions.

\paragraph{2D Joint Regressor.} We use MapReg \cite{chen2022mobrecon} as the 2D joint regressor, which combines the advantages of heatmap- and position-based paradigms. $\mathbf{F}^D_3 \in \mathbb{R}^{(B,C_3,H/16,W/16)}$ and $\mathbf{F}^D_2 \in \mathbb{R}^{(B,C_2,H/8,W/8)}$ are first obtained by a decoder with a skip connection. Next, $\mathbf{F}^D_2$ goes through 2D convolution to adjust the number of channels to match the number of 2D joints. Finally, each feature channel is then flattened into a vector, followed by a multi-layer perceptron (MLP) to generate 2D pose $\mathbf{P}$.

\paragraph{Multi-Scale Pose Feature Extraction.} STMR uses only a single image encoder rather than a stacked structure for running efficiency. In order to improve the pose feature representation and enhance the vertex feature representation, we design the MSPFE module. Figure \ref{fig_mspfe} illustrates the details of the MSPFE module. After obtaining 2D pose $\mathbf{P}$, different scale pose features $\mathbf{F}^P_{E_i}$ and $\mathbf{F}^P_{D_i}$ are obtained with grid sampling \cite{saito2019pifu}. Next, different scale pose features from the visual encoder and decoder are concatenated, followed by MLP to generate multi-scale pose features. The formula of MSPFE is as follows:

\begin{equation}
\begin{aligned}
\mathbf{F}^P &= \operatorname{MLP}([\mathbf{F}^P_{E_0},\dots,\mathbf{F}^P_{E_4},\mathbf{F}^P_{D_3},\mathbf{F}^P_{D_2}]), \\ 
\mathbf{F}^P_{E_i} &= \mathbf{F}^E_i(\mathbf{P}_j)_{j=1,\dots,N}, \quad i = 0,\dots,4, \\
\mathbf{F}^P_{D_i} &= \mathbf{F}^D_i(\mathbf{P}_j)_{j=1,\dots,N}, \quad i = 2,3, \\
\end{aligned}
\label{eq1}
\end{equation}where [$\cdot$] denotes concatenation, and $N$ denotes the number of 2D joints.

\subsection{Vertex Feature Extraction} 
\begin{figure}[t]
\centering
    \subfloat[Predefined lifting matrix. Fingertip-to-vertex lifting matrix is not included in the original MANO model.]{
    \centering
    \includegraphics[width=0.45\textwidth]{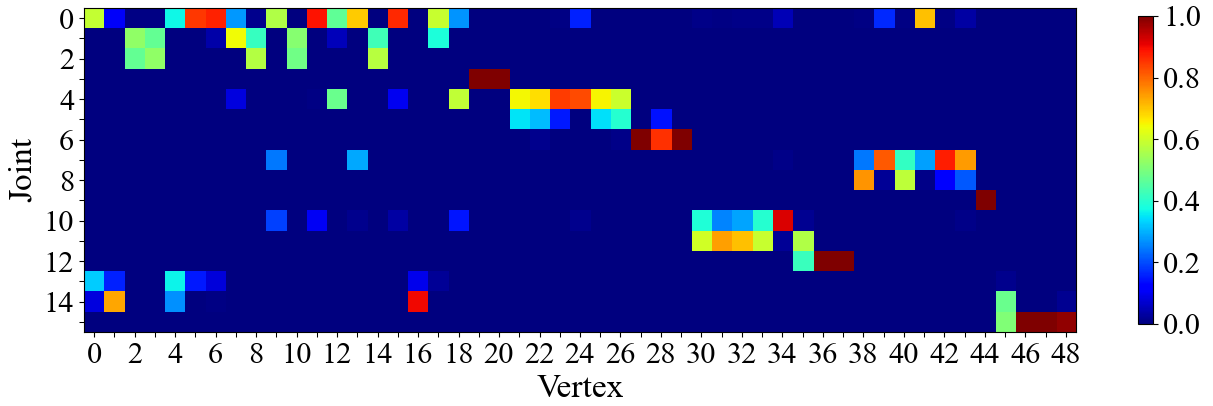}
    \label{fig_ppvl_matrix}
    }
    \hfill
    \subfloat[Positional relationships between poses and vertices. The green dots represent the vertex positions, the blue dots represent the pose positions, and the red dots represent our manually added fingertip positions.]{
    \centering
    \includegraphics[width=0.45\textwidth]{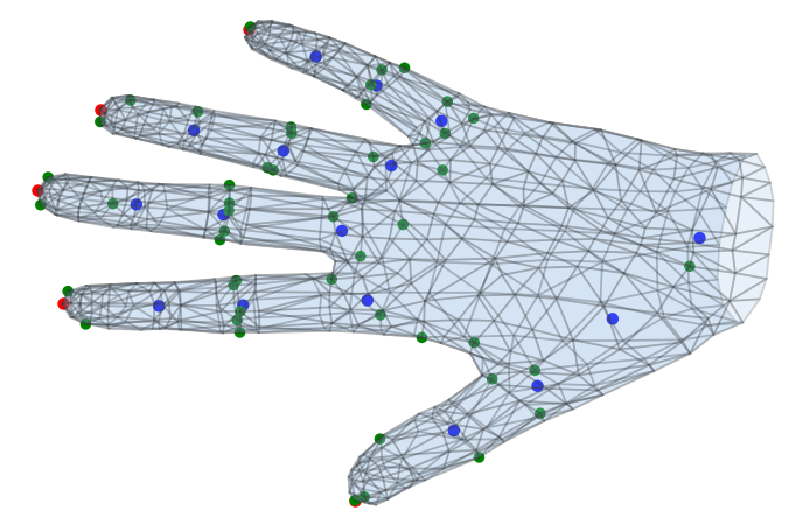}
    \label{fig_ppvl_hand}
    }
\caption{Details of PPVL method. Best viewed in color.}
\label{fig_ppvl}
\end{figure}

After obtaining the pose feature, we design a novel lifting method with a learnable predefined pose-to-vertex matrix, dubbed as PPVL, to obtain the vertex feature. Figure \ref{fig_ppvl} illustrates the details of the PPVL method. MANO-style hand mesh model \cite{romero2022embodied} comprises $V$ vertices and $N$ joints, where $V = 778$, $N = 21$. Following previous works \cite{chen2022mobrecon}, we downsample the template mesh four times by a factor of 2 \cite{garland1997surface} to generate a compact hand mesh with $V_0 = 49$ vertices. The effective transform of pose features $\mathbf{F}^P$ to $V_0$ vertex features is a pivotal factor in obtaining precise hand mesh vertex position.

MANO provides a skin weight between the mesh vertex and the joint, dubbed as $\mathbf{M} \in \mathbb{R}^{(778,16)}$. This weight encapsulates their correlation, enabling precise control over the deformation of the mesh. By sampling $\mathbf{M}$ according to $V_0$, we can obtain the matrix $\mathbf{\hat{M}}_0 \in \mathbb{R}^{(49,16)}$ between the joints and the compact hand mesh with $V_0 = 49$ vertices as shown in Figure \ref{fig_ppvl_matrix}. When the corresponding weight is greater than $0.2$, the mesh vertex is considered part of the joint. In addition, we supplement the lifting matrix $\mathbf{M}_0 \in \mathbb{R}^{(49,5)}$ between the fingertip and $V_0$ mesh vertices. We associate the fingertip with its neighbor mesh vertices, as shown in Figure \ref{fig_ppvl_hand}. In the matrix $\mathbf{M}_0$, we set the value between the fingertip and the neighbor mesh vertices to $0.2$ and set the rest of the values to zero. Finally, $\mathbf{\hat{M}}_0$ and $\mathbf{M}_0$ together form the complete predefined pose-to-vertex lifting matrix $\mathbf{\hat{M}} \in \mathbb{R}^{(49,21)}$ as learnable parameters. Next, vertex features $\mathbf{F}^V_0$ can be calculated as follows:

\begin{equation}
\begin{array}{l}
\mathbf{F}^V_0 = \mathbf{\hat{M}}\mathbf{F}^P.  \\
\end{array}
\label{eq2}
\end{equation}Compared with PVL \cite{chen2022mobrecon}, PPVL contains more prior knowledge and can transform pose features to vertex features more effectively.

\subsection{Mesh Regressor}

Following the extraction of vertex features, our proposed Spiral Transformer is employed to predict the hand mesh vertex positions. We utilize a sparse-to-dense upsampling strategy, where we progressively increase the number of mesh vertices in the order of [49, 98, 195, 389, 778]. This process is facilitated by an upsampling network built upon Spiral Transformer blocks, which not only expands the vertex number but also adaptively adjusts channel number through MLP layers.

\begin{figure}[t]
\centering
    \subfloat[Spiral Neighbor Sampling.]{
    \centering
    \includegraphics[width=0.44\textwidth]{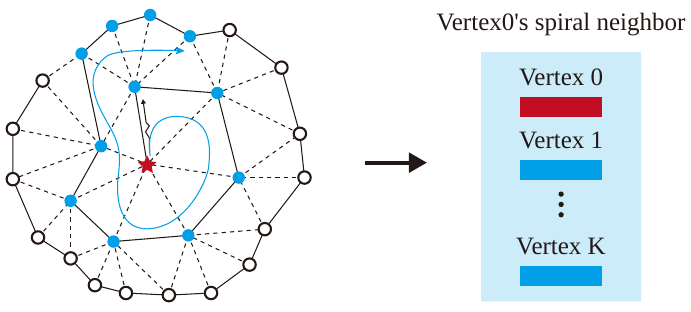}
    \label{fig_sns}
    }
    \hfill
    \subfloat[Spiral Transformer Block.]{
    \centering
    \includegraphics[width=0.44\textwidth]{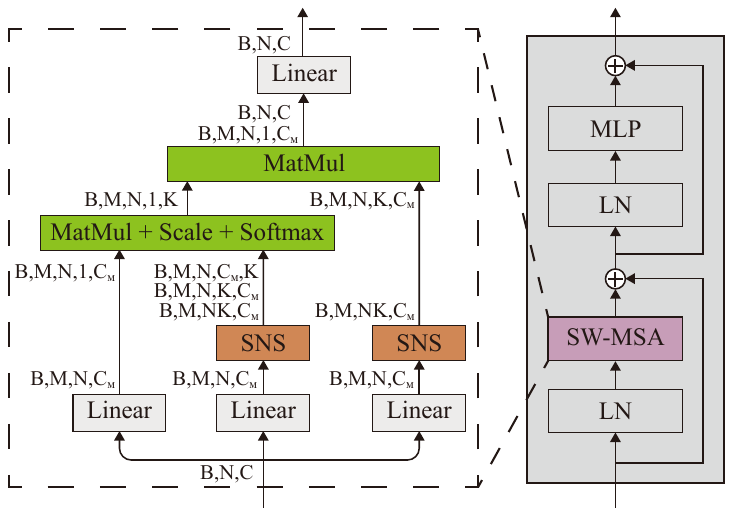}
    \label{fig_stb}
    }
\caption{Details of Spiral Transformer Block. Best viewed in color.}
\label{fig_st}
\end{figure}

As illustrated in Figure \ref{fig_st}, a Spiral Transformer block consists of a spiral-window multi-head self-attention (SW-MSA) module, followed by a 2-layer MLP with GELU non-linearity in the middle. A LayerNorm (LN) layer is applied before each SW-MSA module and each MLP, and a residual connection is applied after each module. The Spiral Transformer block is computed as:

\begin{equation}
\begin{array}{l}
\hat{z}^l = \operatorname{SW-MSA}(\operatorname{LN}(z^{l-1})) + z^{l-1},  \\
{z}^l = \operatorname{MLP}(\operatorname{LN}(\hat{z}^{l})) + \hat{z}^{l},  \\
\end{array}
\label{eq3}
\end{equation}where $\hat{z}^{l-1}$ and $z^{l-1}$ denote the output features of the SW-MSA module and the MLP module for block $l$, respectively

Spiral neighbor sampling method captures the spiral serialization properties of neighbor vertices \cite{gong2019spiralnet++}, which designs a spiral neighbor as:

\begin{equation}
\begin{aligned}
0\operatorname{-ring}(\mathbf{v}) & =\{\mathbf{v}\} \\
(h+1)\operatorname{-ring}(\mathbf{v}) & =\mathcal{N}(h\operatorname{-ring}(\mathbf{v})) \backslash h\operatorname{-disk}(\mathbf{v}) \\
h\operatorname{-disk}(\mathbf{v}) & =\cup_{i=0, . ., h} i\operatorname{-ring}(\mathbf{v}),
\end{aligned}
\label{eq4}
\end{equation}where $\mathcal{N}(\mathbf{V})$ is the set of all vertices adjacent to any vertex in the set $\mathbf{V}$. By explicitly formulating the aggregation order of neighbor vertices, we propose a novel SW-MSA that employs fixed-size neighbors with $K$ vertices and fuses these features.

\begin{table*}[t]
\centering
\caption{Quantitative evaluation. Comparisons with other SOTA methods on the FreiHAND test set. We use \textbf{bold font} to indicate the best performance and use the \underline{underlined font} to represent the second-best performance.}
\begin{tabular}{ccccccc}
\toprule
Method                          & Backbone        & PJ↓             & PV↓             & F@5↑                & F@15↑                & FPS↑              \\
MANO CNN \cite{zimmermann2019freihand} (ICCV 19)              & ResNet50        & 10.9            & 11.0            & 0.516               & 0.934                & -                 \\
Hasson \emph{et al.} \cite{hasson2019learning} (CVPR 19)  & ResNet18        & 13.3            & 13.3            & 0.429               & 0.907                & 20                \\
Boukh \emph{et al.} \cite{boukhayma20193d} (CVPR 19)   & ResNet34        & 13.2            & 35.0            & 0.427               & 0.894                & 11                \\
Kulon \emph{et al.}\cite{kulon2020weakly} (CVPR 20)   & ResNet50        & 8.4             & 8.6             & 0.614               & 0.966                & 60                \\
I2L-MeshNet \cite{moon2020i2l} (ECCV 20)           & ResNet50        & 7.4             & 7.6             & 0.681               & 0.973                & 33                \\
I2UV-HandNet \cite{chen2021i2uv} (ICCV 21)          & ResNet50        & 7.2             & 7.4             & 0.682               & 0.973                & -                 \\
HandAR \cite{tang2021towards} (ICCV 21)                & ResNet50        & 6.7             & 6.7             & 0.724               & 0.981                & 39                \\
CycleHand \cite{gao2022cyclehand} (ACM MM 22)           & ResNet50        & 8.3             & 8.3             & 0.631               & 0.967                & -                 \\
CLIP-Hand3D \cite{guo2023clip} (ACM MM 23)         & ResNet50        & 6.6             & 6.7             & 0.728               & 0.981                & 77                \\
\midrule
                                & ResNet18        & 6.6             & 6.7             & 0.733               & 0.979                & \textbf{106}      \\
                                & ResNet34        & \underline{6.3} & \underline{6.4} & \underline{0.749}   & \underline{0.982}    & \underline{82}    \\
\multirow{-3}{*}{Ours}          & ResNet50        & \textbf{6.2}    & \textbf{6.3}    & \textbf{0.752}      & \textbf{0.984}       & 54                \\
\midrule
\midrule
Pose2Mesh \cite{choi2020pose2mesh} (ECCV 20)             & HRNet+Linear    & 7.4             & 7.6             & 0.683               & 0.973                & 22                \\
MANO GCN \cite{wu2021capturing} (ICME 21)              & HRNet-w48       & 9.5             & 9.5             & 0.579               & 0.950                & -                 \\
CMR \cite{chen2021camera} (CVPR 21)                   & Stack-ResNet50  & 6.9             & 7.0             & 0.715               & 0.977                & \underline{30}    \\
METRO \cite{lin2021end} (ICCV 21)                 & HRNet-w48       & 6.7             & 6.8             & 0.717               & 0.981                & 4                 \\
HIU \cite{zhang2021hand} (ICCV 21)                   & Stack-Hourglass & 7.1             & 7.3             & 0.699               & 0.974                & 9                 \\
MobRecon \cite{chen2022mobrecon} (CVPR 22)              & Stack-ResNet50  & \underline{6.1} & \underline{6.2} & \underline{0.760}   & \underline{0.984}    & \textbf{45}       \\
Fast-METRO \cite{cho2022cross} (ECCV 22)            & HRNet-w48       & 6.5             & -               & -                   & 0.982                & 14                \\
MeshGraphormer \cite{lin2021mesh} (ICCV 21)        & HRNet-w48       & \textbf{5.9}    & \textbf{6.0}    & \textbf{0.765}      & \textbf{0.987}       & 4                 \\
\bottomrule
\end{tabular}
\label{tab_sota}
\end{table*}

Following \cite{vaswani2017attention}, we use positional encodings $\mathbf{P}$ with sine and cosine functions of different frequencies to introduce the sequence order of input $\mathbf{X}$ as follows:

\begin{equation}
\begin{aligned}
\mathbf{\hat{X}} = \mathbf{X} + \mathbf{P}.
\end{aligned}
\label{eq5}
\end{equation}Next, query $\mathbf{Q}$, key $\mathbf{K}$, and value $\mathbf{V}$ can be obtained by using three sets of projections to the input, each consisting of $M$ linear layers (i.e., heads) that map the $C$ dimensional input into a $C_M$ dimensional space, where $C_M = C/M$ is the head dimension. To perform spiral window self-attention, we first need to extract the surrounding spiral neighbor tokens for each query token in the hand mesh. For $i$-th query $\mathbf{Q}_i \in \mathbb{R}^{(B,1,C_M)}$ in each head, keys $\mathbf{K}_i \in \mathbb{R}^{(B,K,C_M)}$ and values $\mathbf{V}_i \in \mathbb{R}^{(B,K,C_M)}$ can be obtained by employing spiral neighbor sampling method (abbreviated as SNS in figure \ref{fig_stb}) as we mentioned earlier. Then, spiral window self-attention for $\mathbf{Q}_i$ can be computed as follows:

\begin{equation}
\begin{aligned}
\operatorname{Attention}(\mathbf{Q}_i, \mathbf{K}_i, \mathbf{V}_i) = \operatorname{Softmax}(\mathbf{Q}_i \mathbf{K}^T_i / \sqrt{C_M}) \mathbf{V}_i.
\end{aligned}
\label{eq6}
\end{equation}The final output is derived by concatenating the output values of each head, followed by a linear projection.

\subsection{Loss Functions}
\paragraph{Vertex Loss \& Pose Loss.} We use $L_1$ norm for 3D mesh loss $\mathcal{L}_{mesh3D}$ and 2D pose loss $\mathcal{L}_{pose2D}$ as follows:

\begin{equation}
\begin{aligned}
\mathcal{L}_{mesh} &= \left\|\mathbf{V}-\mathbf{V}^{gt}\right\|_1, \\
\mathcal{L}_{pose2D} &= \left\|{\mathbf{P}}-{\mathbf{P}}^{gt}\right\|_1,
\end{aligned}
\label{eq7}
\end{equation}where $\mathbf{V}$ is vertex sets of a mesh; the superscript $gt$ denotes the ground true. 

\paragraph{Mesh Smooth Loss.} To ensure the geometric smoothness of the predicted vertices, normal loss $\mathcal{L}_{norm}$ and edge length loss $\mathcal{L}_{edge}$ are adopted:

\begin{equation}
\begin{aligned}
\mathcal{L}_{norm} &= \sum_{\mathbf{c} \in \mathbf{C}} \sum_{(i, j) \subset \mathbf{c}}\left|\frac{\mathbf{V}_i-\mathbf{V}_j}{\left\|\mathbf{V}_i-\mathbf{V}_j\right\|_2} \cdot \mathbf{n}_{\mathbf{c}}^{gt}\right|, \\
\mathcal{L}_{edge} &= \sum_{\mathbf{c} \in \mathbf{C}} \sum_{(i, j) \subset \mathbf{c}}\left|\left\|\mathbf{V}_i-\mathbf{V}_j\left\|_2-\right\| \mathbf{V}_i^{gt}-\mathbf{V}_j^{gt} \|_2\right|\right.,
\end{aligned}
\label{eq8}
\end{equation}where $\mathbf{C}$ is triangular face sets of a mesh; $\mathbf{n}_{\mathbf{c}}^{gt}$ denotes unit normal vector of triangular face $\mathbf{c}$.

\paragraph{Consistency loss.} Following several self-supervision related works, consistency supervision in both 3D and 2D space are adopted:

\begin{equation}
\begin{aligned}
\mathcal{L}_{con3D} & =\left\|R_{1 \rightarrow 2} \mathbf{V}_{view1}-\mathbf{V}_{view2}\right\|_1, \\
\mathcal{L}_{con2D} & =\left\|T_{1 \rightarrow 2} \mathbf{P}_{view1}-\mathbf{P}_{view2}\right\|_1,
\end{aligned}
\label{eq9}
\end{equation}where $R_{1 \rightarrow 2}$ is the relative rotation between two views; $T_{1 \rightarrow 2}$ is the relative affine transformation between two views.

Finally, STMR is trained in an end-to-end manner with total loss $\mathcal{L}_{total} = \mathcal{L}_{mesh} + \mathcal{L}_{pose2D} + \alpha_{n} \mathcal{L}_{norm} + \alpha_{e} \mathcal{L}_{edge} + \mathcal{L}_{con3D} + \mathcal{L}_{con2D}$, where $\alpha_{e} = 0.5, \alpha_{n} = 0.05$ are used to balance different loss terms. 

\section{Experiments}

\begin{figure*}[th]
\centering
    \subfloat[Normal poses.]{
    \centering
    \includegraphics[width=0.90\textwidth]{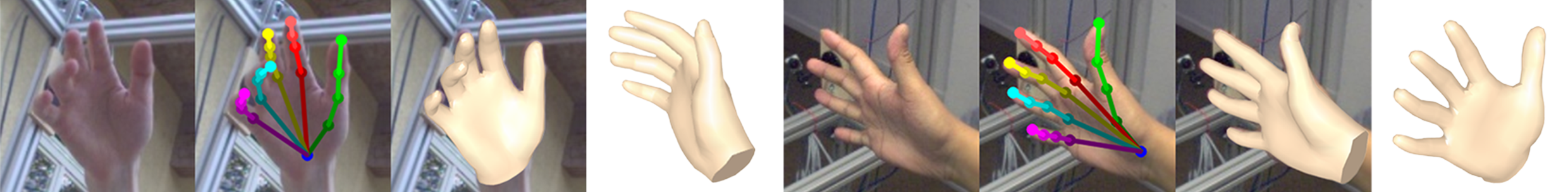}
    \label{fig_nv}
    }
    \hfill
    \subfloat[Challenging poses.]{
    \centering
    \includegraphics[width=0.90\textwidth]{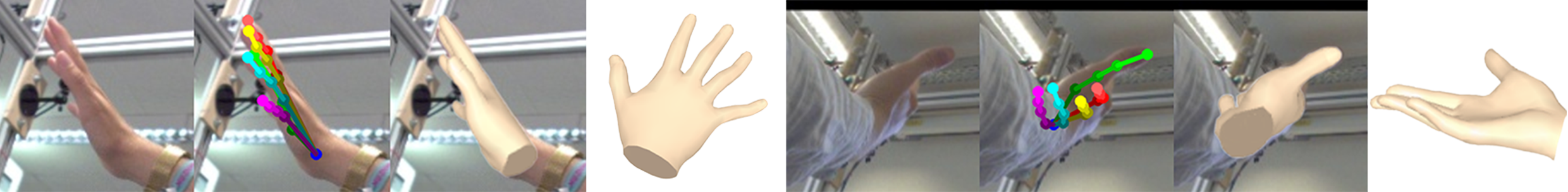}
    \label{fig_cp}
    }
    \hfill
    \subfloat[Object occlusion.]{
    \centering
    \includegraphics[width=0.90\textwidth]{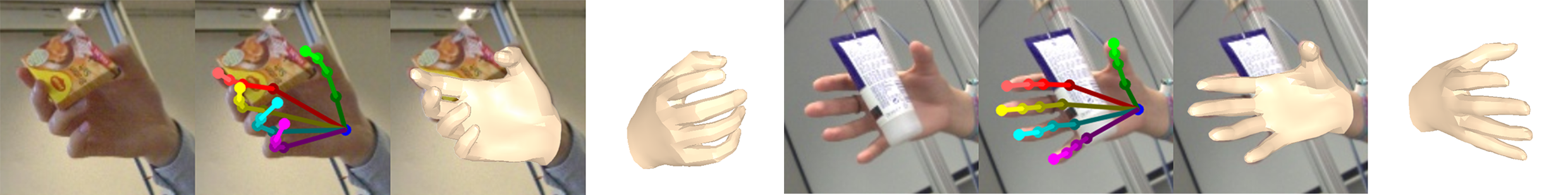}
    \label{fig_oc}
    }
\caption{Qualitative results of the proposed STMR. We show six input images, with their 2D pose, projection of mesh, and side-view mesh. Best viewed in color.}
\label{fig_qualitative}
\end{figure*}

\begin{table}[h]
\centering
\caption{Ablation study of MSPFE module and PPVL. A check mark (\checkmark) denotes that the module is used.}
\begin{tabular}{ccccc}
\toprule
MSPFE      & PPVL       & 3D AUC(0-50mm)↑ & PJ↓  & PV↓  \\
\midrule
           &            & 0.861           & 6.89 & 6.98 \\
           & \checkmark & 0.863           & 6.81 & 6.90 \\
\checkmark &            & 0.865           & 6.71 & 6.81 \\
\checkmark & \checkmark & 0.866           & 6.61 & 6.72 \\
\bottomrule
\end{tabular}
\label{tab_ab_module}
\end{table}

\begin{table}[h]
\centering
\caption{Ablation study of 3D decoding.}
\begin{tabular}{ccccc}
\toprule
Mesh Regressor & 3D AUC(0-50mm)↑ & PJ↓   & PV↓  \\
\midrule
SpiralConv++   & 0.864           & 6.75  & 6.86 \\
DSConv         & 0.863           & 6.79  & 6.91 \\
SW-MSA         & 0.866           & 6.61  & 6.72 \\
Global         & 0.866           & 6.65  & 6.74 \\
\bottomrule
\end{tabular}
\label{tab_ab_mesh}
\end{table}

\subsection{Datasets \& Metrics}

We conduct experiments on commonly-used hand mesh reconstruction benchmarks as listed below.

\paragraph{FreiHand.} The FreiHand dataset \cite{zimmermann2019freihand} contains 130,240 training images and 3,960 evaluation samples. It is collected from a diverse group of 32 individuals, featuring varying genders and ethnic backgrounds, ensuring a broad representation for research purposes.

We use the following metrics to quantitatively evaluate model performance.

\paragraph{MPJPE \& MPVPE.} MPJPE and MPVPE are quantitative metrics that evaluate the accuracy of joint and vertex position predictions by calculating the average Euclidean distance (in millimeters) between the estimated and ground-truth coordinates.

\paragraph{PA-MPJPE \& PA-MPVPE.} PA-MPJPE/MPVPE (abbreviated as PJ/PV) represents a variant of the original MPJPE/MPVPE, which incorporates Procrustes analysis to address global variations \cite{gower1975generalized}.

\paragraph{AUC.} AUC is the area under the curve of PCK (percentage of correct keypoints) vs. error thresholds.

\paragraph{F-Score.} F-Score is the harmonic mean between recall and precision between two meshes with respect to a specific distance threshold. F@5 and F@15 denote the F-Score at distance thresholds of 5mm and 15mm, respectively.

\subsection{Implementation Details}
We use the ImageNet pre-trained weight parameters to initialize the image encoder and train STMR in an end-to-end manner with an Adam optimizer for a total of 48 epochs. The initial learning rate is set to $10^{-3}$. After 38 epochs, the learning rate is divided by 10. The mini-batch size is set to 32 for training. The number of spiral neighbor vertices $K$ is set to 9. During the training and inference stages, we use PyTorch to conduct all experiments. We train the network on a single NVIDIA RTX 3090 and test the network on a single NVIDIA RTX 2080Ti.

\begin{figure*}[h]
\centering
\includegraphics[scale=0.23]{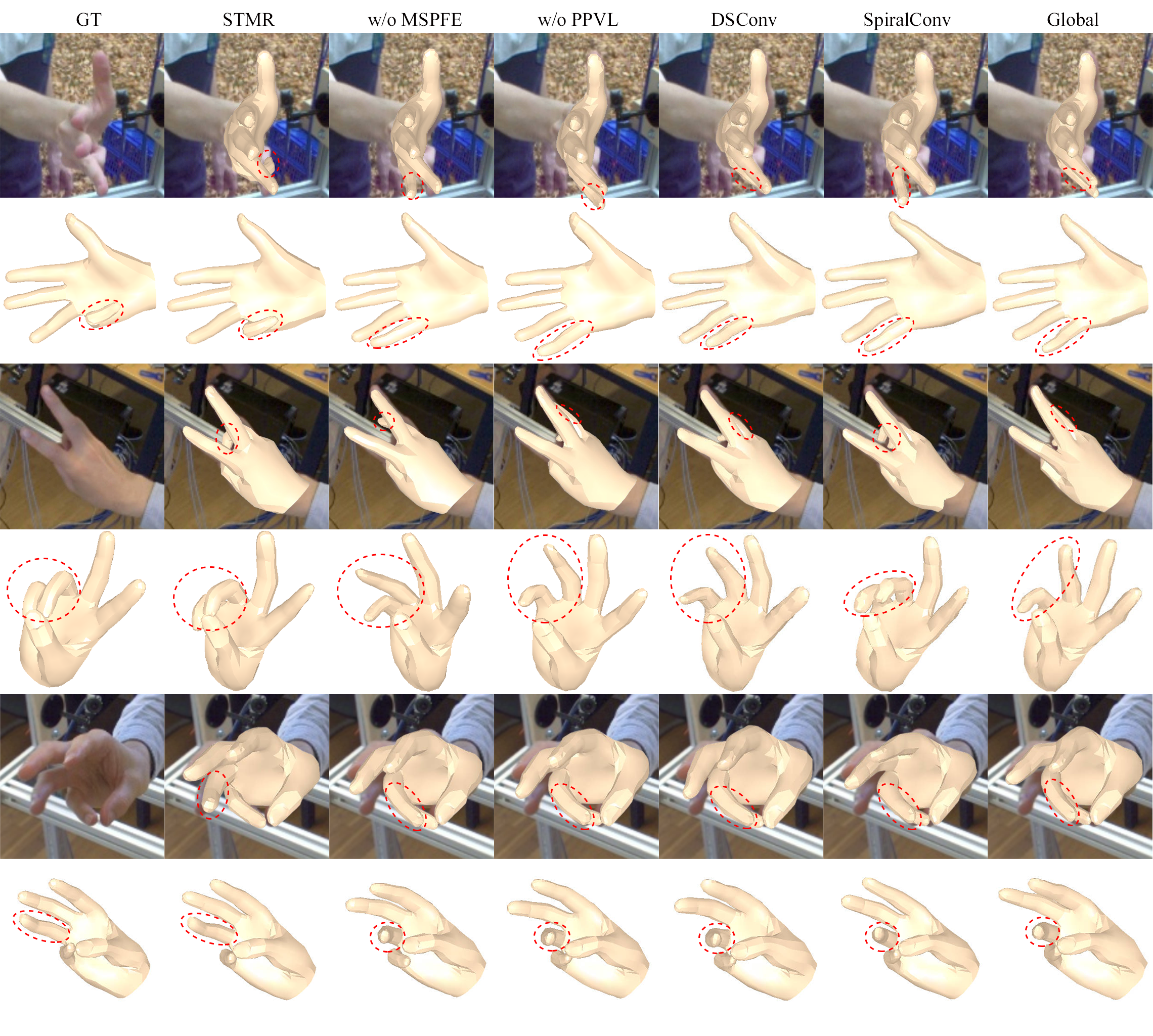}
\caption{Ablation study. We show results in camera view and side view respectively. We exclusively provide ground truth (GT) annotations for the side view perspective. Best viewed in color.}
\label{fig_ab}
\end{figure*}

\subsection{Quantitative and Qualitative Results}
\paragraph{Quantitative Results.}
For the FreiHAND dataset, we conduct a detailed comparison primarily with the current state-of-the-art (SOTA) methods. As shown in Table \ref{tab_sota}, our method achieves SOTA performance and inference speed among methods employing a single ResNet as the visual encoder. Compared with the best CLIP-Hand3D, STMR achieves comparable performance (PJ: 6.6 vs. \textbf{6.6}, PV: 6.7 vs. \textbf{6.7}, F@5: 0.728 vs. \textbf{0.733}) with ResNet18 while maintaining a significant speed advantage (FPS: 77 vs. \textbf{106}). As the visual encoder model scales up, our method achieves SOTA accuracy. Utilizing ResNet34 as an image encoder, STMR achieves significant performance (PJ: 6.6 vs. \textbf{6.3}, PV: 6.7 vs. \textbf{6.4}, F@5: 0.728 vs. \textbf{0.749}) and slight speed improvements (FPS: 77 vs. \textbf{82}). Furthermore, our approach still achieves fast inference (FPS: 4, 45 vs. \textbf{54}) and competitive performance (PJ: 5.9, 6.1 vs. \textbf{6.2}, PV: 6.0, 6.2 vs. \textbf{6.3}, F@5: 0.765, 0.760 vs. \textbf{0.752}) compared with SOTA models using large-scale visual encoders (MeshGraphormer and MobRecon). Note that the performance of some methods using supplementary or mixed datasets is not included in our statistics.

\paragraph{Qualitative Results.}
As shown in Figure \ref{fig_qualitative}, we randomly select some test images from the FreiHAND dataset and then reconstruct 3D hand mesh using our method. These images demonstrate a variety of hand poses, including self-occlusion and object occlusion. In normal poses, STMR can accurately reconstruct the hand mesh. Even under extreme view, our method demonstrates strong robustness. When the hand is occluded, STMR can utilize the information of the unoccluded hand part to accurately recover the occluded hand mesh. Overall, our model achieves accurate hand mesh reconstruction even with complex hand images.

\begin{figure*}[h]
\centering
\includegraphics[scale=0.60]{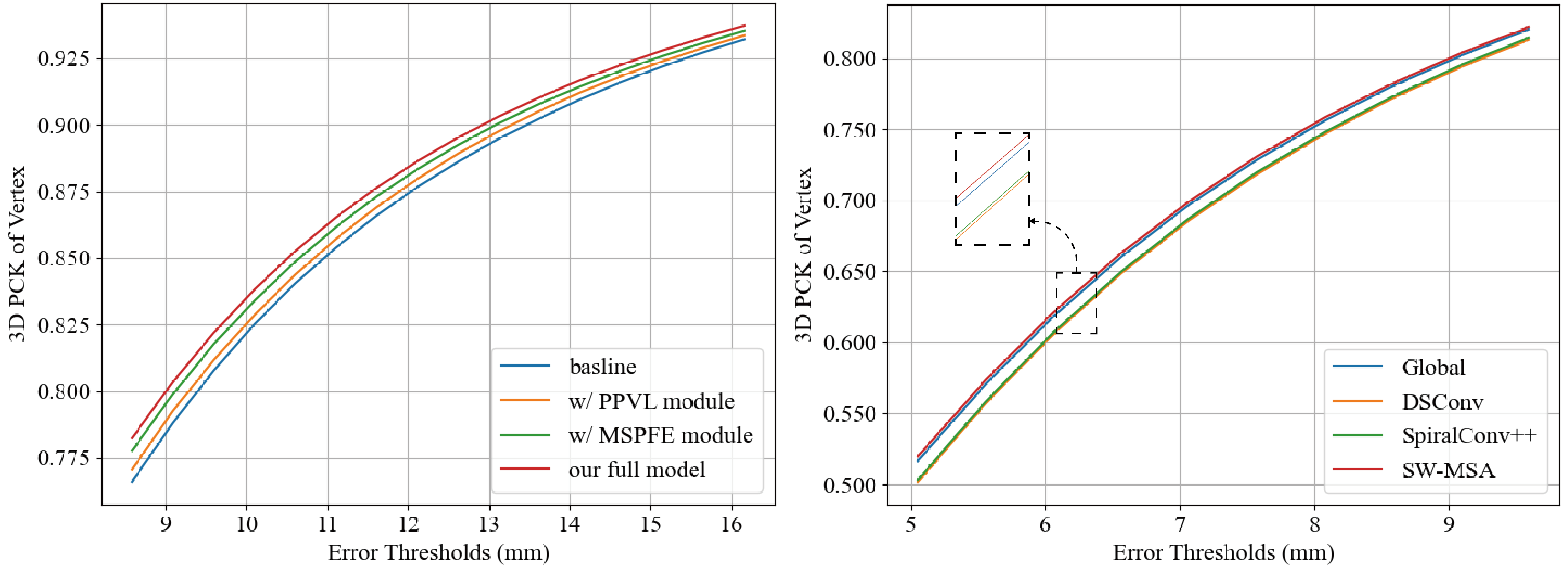}
\caption{3D PCK vs. error thresholds. It measures the performance of different structural configurations on the FreiHAND validation set. Best viewed in color.}
\label{fig_3dpcv}
\end{figure*}

\subsection{Ablation Study}

\paragraph{Ablation Study of MSPFE Moduel and PPVL Method.} 
In Table \ref{tab_ab_module}, we compare the performance of the MSPFE module and the PPVL method on the FreiHand validation set. MSPFE effectively leverages multi-scale features, resulting in significantly improved hand mesh reconstruction accuracy (3D AUC: 0.861 vs. \textbf{0.865}, PJ: 6.89 vs. \textbf{6.71}, PV: 6.89 vs. \textbf{6.81}) compared with single-scale features. We believe that the MSPFE module can provide rich information for subsequent mesh vertex regression, thus facilitating more accurate hand mesh reconstruction. PPVL, on the other hand, utilizes the prior knowledge from the MONO model, outperforming randomly initialized pose-to-vertex matrices (3D AUC: 0.861 vs. \textbf{0.863}, PJ: 6.89 vs. \textbf{6.81}, PV: 6.98 vs. \textbf{6.90}). The PPVL plays a pivotal role in extracting precise vertex features by leveraging prior information for accurate 2D-to-3D mapping. When they are combined, the full model further enhances the reconstruction performance (3D AUC: 0.861 vs. \textbf{0.866}, PJ: 6.89 vs. \textbf{6.61}, PV: 6.98 vs. \textbf{6.72}). In Figure \ref{fig_ab}, we qualitatively compare the inference results with and without the MSPFE module or PPVL on several images, respectively. The details are indicated by the red dashed boxes. Intuitively, the MSPFE module and PPVL help the model to reconstruct a more accurate hand mesh. In addition, we present the 3D PCK curve within a specified range in Figure \ref{fig_3dpcv} to quantitatively illustrate these results.

\paragraph{Ablation Study of 3D Decoding.} 
In Table \ref{tab_ab_mesh}, we compare the model performance of different 3D decoding methods on the FreiHand validation set. Compared with SpiralConv++ and DSConv, SW-MSA significantly improves hand mesh reconstruction (3D AUC: 0.864, 0.863 vs. \textbf{0.866}, PJ: 6.75, 6.79 vs. \textbf{6.61}, PV: 6.86, 6.91 vs. \textbf{6.72}). This result indicates that the Transformers' capacity to model long-term dependencies is particularly crucial when handling one-dimensional vertex features generated through spiral neighbor sampling, as it helps reveal the relationship among vertices. Compared with global self-attention (Global), SW-MSA obtains higher accuracy (3D AUC: 0.886 vs. \textbf{0.866}, PJ: 6.65 vs. \textbf{6.61}, PV: 6.74 vs. \textbf{6.72}) in hand mesh reconstruction, underscoring the topological information provided by spiral neighbor sampling in the Transformer is important. SW-MSA focuses solely on local neighbors, achieving superior reconstruction performance compared with global inter-vertex interactions. This suggests that relying on positional encoding alone to represent mesh structures is inadequate. In addition, we provide a qualitative comparison of the 3D decoder selection in Figure \ref{fig_ab}, which shows that SW-MSA can reconstruct a more accurate hand mesh. In addition, we present the 3D PCK curve within a specified range in Figure \ref{fig_3dpcv} to quantitatively illustrate these results.

\section{Conclusion}
In this paper, we propose a novel hand mesh reconstruction method with superior efficiency and accuracy. First, we propose an MSPFE module for extracting rich pose features under a single image encoder. Then, a PPVL method is designed to accurately obtain vertex features in 2D-to-3D mapping, which fully utilizes the a priori information in the MANO model. Besides, a Spiral Transformer is developed to handle the 3D decoding task effectively by injecting mesh topology into the Transformer through spiral neighbor sampling. Finally, the rationality and superiority of our method are effectively demonstrated through comprehensive ablation experiments and comparisons with several hand mesh reconstruction methods on the FreiHand dataset. In the future, we plan to investigate an efficient hand texture model.

\section{Acknowledgments}
This work was supported by the National Natural Science Foundation of China under Grant No.62376100 and Grant No.61976095 and the Natural Science Foundation of Guangdong Province of China under Grant No.2022A1515010114.

{\small
\bibliographystyle{ieee_fullname}
\bibliography{egbib} 
}

\end{document}